\documentclass[10pt, a4paper, onecolumn]{dukenlp}

\usepackage{natbib}
\usepackage{subcaption}
\usepackage{float}
\usepackage[most]{tcolorbox}
\usepackage{latexsym}
\usepackage{bbm}
\usepackage{listings}
\usepackage{multirow}
\usepackage{comment}
\usepackage{xspace}
\usepackage{array}
\usepackage{wrapfig}

\definecolor{aback}{HTML}{EFF6FF}
\definecolor{aframe}{HTML}{93C5FD}
\newtcolorbox{PromptBox}[1][]{
  enhanced,
  colback=aback,
  colframe=aframe,
  title=Prompt,
  sharp corners,
  breakable,
  #1
}

\definecolor{darkblue}{rgb}{0, 0, 0.5}

\title{How Much Backtracking is Enough? Exploring the Interplay of SFT and RL in Enhancing LLM Reasoning}

\author{%
Hongyi James Cai\textsuperscript{1,*},
Junlin Wang\textsuperscript{1,*},
Xiaoyin Chen\textsuperscript{2,3},
Bhuwan Dhingra\textsuperscript{1}
\\
\textsuperscript{1}Duke University,
\textsuperscript{2}Mila -- Qu\'ebec AI Institute,
\textsuperscript{3}Universit\'e de Montr\'eal
\\
\texttt{\{hongyi.cai, junlin.wang2\}@duke.edu, xiaoyin.chen@mila.quebec,}
\\
\texttt{bdhingra@cs.duke.edu}
}

\begin{document}

\maketitle
\begingroup
\renewcommand{\thefootnote}{*}
\footnotetext{Equal contribution.}
\endgroup

\begin{abstract}
Recent advancements in large language models (LLMs) suggest that reinforcement learning (RL) effectively internalizes search strategies, yielding significant improvements on challenging reasoning tasks through extended chains of thought. While backtracking is widely viewed as the core mechanism enabling this improvement, its precise dynamics and how SFT and RL interact to optimize for it, remain poorly understood. In this work, we argue that backtracking—the ability to explicitly revisit and revise earlier reasoning steps—is a key operator that enables the transition from linear generation to non-linear tree search for LLMs, unlocking strong performance on combinatorially complex tasks.
We motivate this claim by first studying the interplay between supervised fine-tuning (SFT) warm-up and RL across eight reasoning tasks: Countdown, Sudoku, Arc 1D, Advanced Geometry, Color Cube Rotation, List Functions, Zebra Puzzles, and Self Reference. We find that a regular SFT warm-up using self-sampled CoT sequences provides a modest boost to subsequent RL compared to training without any SFT warm-up; however, such gains saturate as tasks become more difficult.
Motivated by this failure mode, we introduce a backtracking-centric training recipe. By synthetically varying the number of explicit backtracking steps in the SFT warm-up, we show that (i) longer CoTs containing backtracks stabilize and amplify RL, and (ii) the optimal backtrack depth scales with task difficulty—zero for Arc 1D, one for Countdown, and five or more for Sudoku. Our findings establish that principled backtracking is essential for improving the model's reasoning capability.
\end{abstract}

\section{Introduction}

\begin{figure*}
  \centering
  \includegraphics[width=\textwidth]{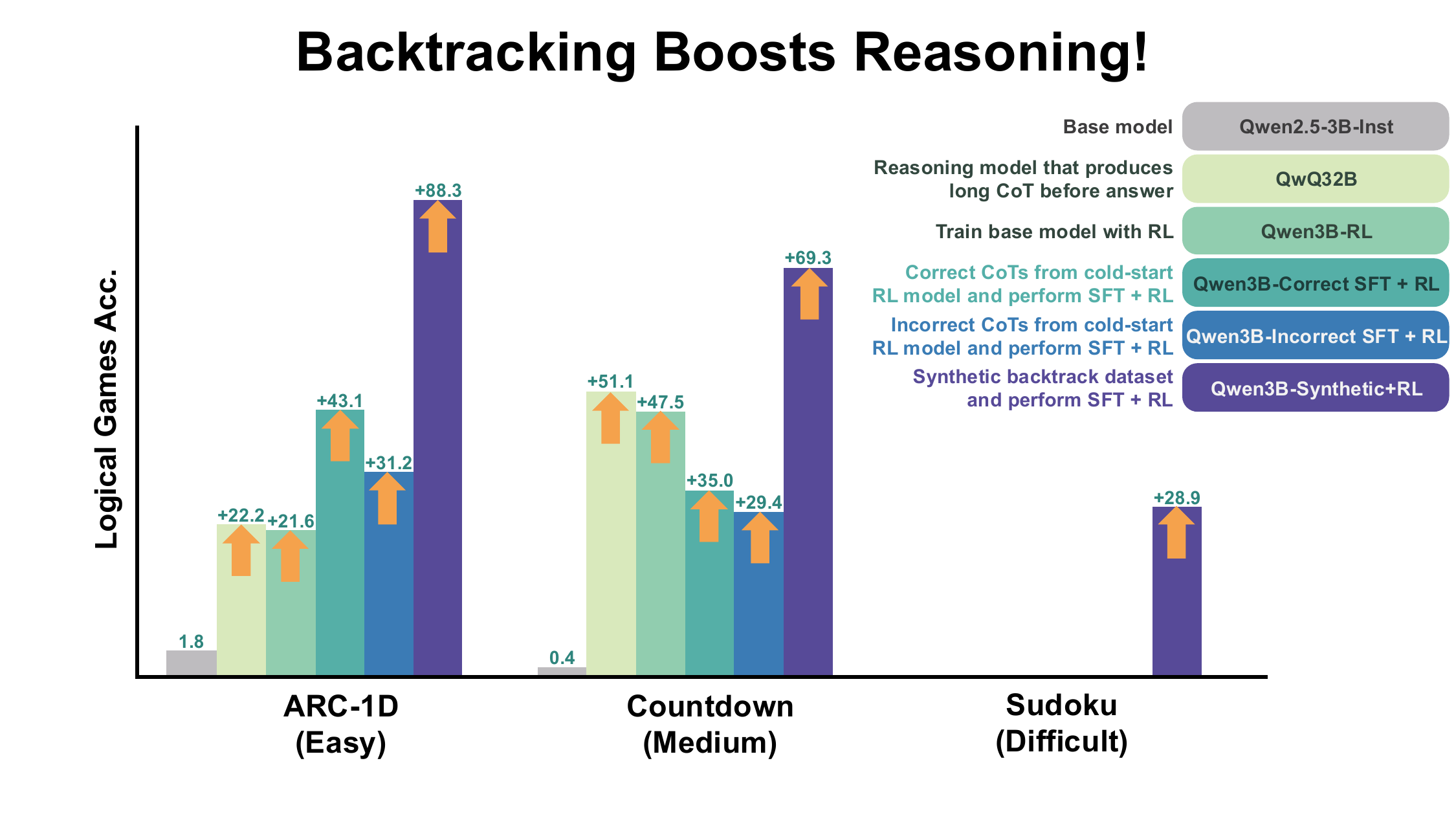}
  \caption{We engineer SFT data with synthetic backtracks to achieve significant improvements over three reasoning tasks, Arc-1D, Countdown, and Sudoku, each representing a different difficulty.}
  \label{fig:illustration}
\end{figure*}

    Recent large language models such as DeepSeek-R1 \citep{guo2025deepseek} and OpenAI's o1 \citep{o1} have demonstrated remarkable reasoning abilities on various complex tasks. Their success stems from allocating substantially more inference‑time compute, letting the model perform an internal search via extended chains of thought before committing to an answer \citep{snell2024scaling}. Reinforcement learning has emerged as the most effective way to unlock these long reasoning traces \citep{shao2024deepseekmath, team2025kimi, schulman2017proximal, zelikman2022star, rafailov2023direct}. In the DeepSeek‑R1 era, a key shift is from open‑ended process rewards that score intermediate steps \citep{lightman2024let, guan2025rstar,zhang2025lessons,zheng2025processbench} to verifiable, rule‑based rewards that check final answers \citep{zeng2025simplerl,chen2025empirical}, providing a sharper signal that steers exploration toward correct trajectories.

     Although reinforcement learning often improves LLM reasoning, the scope and limits of the underlying enhancement mechanisms remain under-explored. \citet{gandhi2025cognitive} highlight this by showing large RL gains for Qwen but only marginal ones for Llama. One hypothesis attributes Qwen’s advantage to the use of “thinking” tokens such as “wait” and “verify” \citep{xie2025logic, gandhi2025cognitive}. More broadly, RL appears to primarily amplify behaviors already present in the base model---either inherited from pretraining \citep{zhao2025echo} or generalized from SFT warm-up \citep{chu2025sft}---rather than consistently improving accuracy through longer traces alone \citep{qin2025backtrack, wang2025think, yue2025does}.
    
    This motivates a closer look at backtracking, which increasingly emerges in top reasoning models such as DeepSeek-R1 \citep{guo2025deepseek} and OpenAI's o1 series \citep{o1}, and suggests a shift from \emph{reasoning-as-generation} to \emph{reasoning-as-search}. In classical cognitive science, problem solving is often framed as search through a state space \citep{newell1972human}; for autoregressive LLMs, backtracking is the key operator that breaks strict linear generation by treating prior outputs as revisable state assignments rather than immutable facts. Viewed through this lens, backtracking is not a task-specific trick but a general operator for tree search---central to exploration, planning, and multi-hop reasoning \citep{qin2025backtrack}. It ties together detecting a flawed partial solution (verification) and revising it (recovery), yielding a concrete, controllable handle on search behavior. This perspective motivates our study of how SFT warm-up interacts with RL via backtracking frequency: when is warm-up necessary, which warm-ups matter, and how much backtracking is optimal before it becomes counter-productive?
    
    To answer these questions, we run controlled experiments that isolate how the \emph{presence} and \emph{structure} of warm-up data shape downstream RL. We first benchmark pure RL (no reasoning specific SFT warm-up) against cold-start RL using self-sampled CoTs as warm-up, showing that simple warm-ups can help but become insufficient on harder tasks. We then introduce synthetic warm-ups with explicit, tunable backtracking (constructed via depth-first search or heuristic search for three representative tasks) and show that backtracking is a reliable knob for unlocking RL. 
    
    Taken together, our controlled experiments reveal a precise recipe for effective RL: initializing with synthetic SFT data containing an appropriate number of backtracks—matched to the difficulty of the task—consistently pushes the model beyond its base reasoning capabilities.

Our main contributions are:
    \begin{itemize}
        \item A controlled comparison of pure RL vs. cold-start RL across eight reasoning tasks, clarifying when standard self-sampled CoT warm-ups help and when they plateau (Section~\ref{sec:control-content-vary-structure}).
        \item A backtracking-centric SFT+RL recipe based on synthetic trajectories with calibrated backtrack depth, which substantially improves performance on combinatorially hard problems (Section~\ref{sec:pushing-boundries-backtrack}).
        \item Additional analysis showing that RL can still benefit from format-aligned but semantically incorrect CoT warm-ups (Section~\ref{sec:additional-analysis}).
    \end{itemize}

\section{Methodology}
    In this study, we conduct a comprehensive analysis using controlled experiments to investigate the interplay between SFT and RL, focusing on how different training data mixtures influence RL outcomes. Specifically, we examine which types of SFT warm-ups benefit RL and seek to understand the underlying reasons. Table~\ref{tab:SFT_setups} presents the SFT setups we evaluate: no SFT, self-sampled SFT, synthetic backtracking SFT, and shuffled SFT. Ultimately, we aim to equip the model with the ability to tackle difficult problems beyond its base capabilities.

\begin{table}[t]
  \caption{Brief explanation of each SFT warm-up setup explored in this paper.}
  \label{tab:SFT_setups}
  \begin{center}
    \begin{small}
      \begin{sc}
        \begin{tabular}{ll}
          \toprule
          \textbf{Abbr.} & \textbf{Explanation} \\
          \midrule
          No-SFT &
          RL with no SFT (Pure RL) \\
          Self-sampled SFT &
          SFT on model-generated trajectories \\
          Synthetic backtrack SFT &
          SFT on synthetic trajectories with backtracking \\
          Shuffled SFT &
          SFT on trajectories shuffled across problems \\
          \bottomrule
        \end{tabular}
      \end{sc}
    \end{small}
  \end{center}
  \vskip -0.1in
\end{table}

We employ a total of 8 reasoning tasks, adopted from Reasoning Gym \citep{stojanovski2025reasoning}, and select a subset to further curate synthetic data to ensure our experiments provide analytical insights. 

\subsection{Why backtracking matters}
\label{sec:backtracking-importance}
In the context of LLM reasoning, we define backtracking as a formal state-space traversal operator. Let a reasoning task be defined by a state space $S$ and a set of transitions $T$. A standard chain-of-thought approach attempts to produce a path $P = (s_0, s_1, \ldots, s_n)$ such that $s_n$ is a goal state. For complex tasks, the probability that every local transition $p(s_{i+1} \mid s_i)$ is correct can be low, implying a cumulative success probability that decays quickly with path length.

We frame backtracking as a transition $T_{\mathrm{back}}: s_i \to s_j$ where $j < i$. This operator serves two key cognitive functions: (i) \textbf{validation-driven pruning}, where the model evaluates a partial state $s_i$ against task constraints and prunes a branch upon detecting an inevitable dead-end; and (ii) \textbf{state recovery}, where the model restores a previous valid state $s_j$ to explore an alternative transition $s_j \to s'_{j+1}$. By varying the backtrack depth in our synthetic SFT data, we effectively condition the model to recognize its search frontier and dynamically allocate computation during inference to explore alternative paths. Our framework allows backtracking frequency to be treated as a hyperparameter for Search-Aware SFT.

Finally, many common test-time scaling strategies can be viewed as implicit forms of backtracking: \textbf{parallel sampling} explores alternative branches by generating multiple candidate continuations (i.e., shallow backtracks across independent rollouts), and \textbf{self-reflection} provides the validation signal that a current trajectory is flawed \citep{kim2025astro}. In both cases, verification or reflection can indicate that a current path is incorrect, but explicit backtracking is the recovery action that returns to a prior state to pursue a different transition.

\subsection{Reasoning Games}
The puzzle-style reasoning tasks we choose in this study encapsulates many essential properties of reasoning: multi-step deductive reasoning, symbolic manipulation under constraints, search over discrete combinatorial structures, and systematic backtracking when partial solutions fail. These are the same algorithmic primitives that underlie mathematical proofs and algebraic transformation, enabling our study to provide a proxy for broader scope of tasks. The specific reasoning games chosen are Countdown, Advanced Geometry, Arc-1D, Sudoku, Color Cube Rotation, Zebra Puzzles, List Functions, and Self Reference (Appendix~\ref{appendix:reason-games}).

\subsection{Model and training}
\label{sec:model_and_training}
We use the Qwen2.5 family of models \citep{qwen2.5}, primarily focusing on Qwen2.5-3B-Instruct for our pure-RL and self-sampled-SFT experiments. For our synthetic-backtracking experiments on Arc-1D, Countdown, and Sudoku, we show results with Qwen2.5-3B-Instruct, Qwen2.5-7B-Instruct, and Llama3.2-3B-Instruct \citep{grattafiori2024llama}. We report pass@1 accuracy, evaluating only the correctness of the final answer placed inside \texttt{<answer></answer>} tokens.

For both SFT and RL training, we adopted code from \citet{sheng2025hybridflow, tinyzero}, the specific rollout lengths depend on whether the model is initialized from long or short CoT. Typically, for RL training initialized from long CoTs, we use rollout length from 4k to 8k tokens, depending on the tasks. For short CoTs, we use rollout length from 1k to 2k tokens, depending on the tasks (Appendix~\ref{appendix:rl-details}).

We adopt rule-based rewards \citep{guo2025deepseek}, where format accuracy (successfully generate a pair of \texttt{<think></think>} and \texttt{<answer></answer>} tokens with think tokens come before answer tokens) is rewarded 0.1 point, and the answer accuracy-exact match with the ground truth-is rewarded 0.9 point (Appendix~\ref{appendix:reward-setup}). Together, a problem may be rewarded a maximum of 1.0 point and the minimal of 0.0 point. 

\section{Maximizing base model reasoning capabilities}
\label{sec:control-content-vary-structure}

To systematically investigate how SFT interacts with reinforcement learning in enhancing large language models' reasoning capabilities, we scientifically address a foundational question: is a warm-up stage necessary for effective RL? Specifically, we begin with two distinct approaches—pure RL (no SFT warm-up) and cold-start RL with a short self-sampled CoT warm-up—to establish baselines and evaluate their efficacy across diverse reasoning tasks. These experiments demonstrate that, although pure RL is already effective, incorporating a simple self-sampled CoT warm-up further improves performance. However, both methods face significant limitations when applied to more challenging reasoning tasks, such as Sudoku.

\subsection{Pure RL - works albeit limited}
\label{pg:no-sft}
We first explored pure RL, a direct RL approach without any reasoning specific SFT warm-up. The motivation behind this was to establish a clear baseline and determine whether the model could independently internalize effective search strategies and concise reasoning through pure RL exposure. In Figure~\ref{fig:short-cots-eval}, we observe that just pure RL itself is able to witness improvements in model's reasoning abilities, compared to their base model. We manually inspect a few of the generated trajectories and find that pure RL enables the model to internalize rudimentary strategies, such as explicit verbalized reasoning (e.g., in Appendix~\ref{appendix:model_output-cd-short}) and concise thinking (e.g., in Appendix~\ref{appendix:model_output-lf-short}). It is noticeable that for most tasks, pure RL naturally produces linear trajectories without backtracking. For Countdown, the model produces backtracking but it is unstructured. Due to probable pre-training data exposure, the model occasionally attempts to correct itself. However, these corrections are stochastic and often "lazy"—retrying the whole calculation rather than pinpointing the error.

Additionally, training the model in one environment also improves performance in other tasks as shown in Appendix~\ref{appendix:full-cold-start-eval}. Training in Advanced Geometry environment induces good generalization to Zebra Puzzles and List Functions; training in Countdown induces generalization to Advanced Geometry. However, limitations quickly became apparent, particularly on combinatorial problems like Sudoku and Zebra Puzzles, indicating that pure RL alone is insufficient for complex reasoning tasks.

\subsection{Self-sampled short SFT as warm-up goes beyond pure RL}
\label{sec:self-sampled SFT}

Building on the insights from pure RL, we hypothesized that a supervised fine-tuning warm-up using short self-sampled CoTs could further enhance reasoning performance by priming models with initial cognitive strategies.

\paragraph{Training and evaluating setup}
\label{sec:short-cots}

\begin{figure}[t]
\centering

\begin{subfigure}[t]{0.48\columnwidth}
    \centering
    \includegraphics[width=\linewidth]{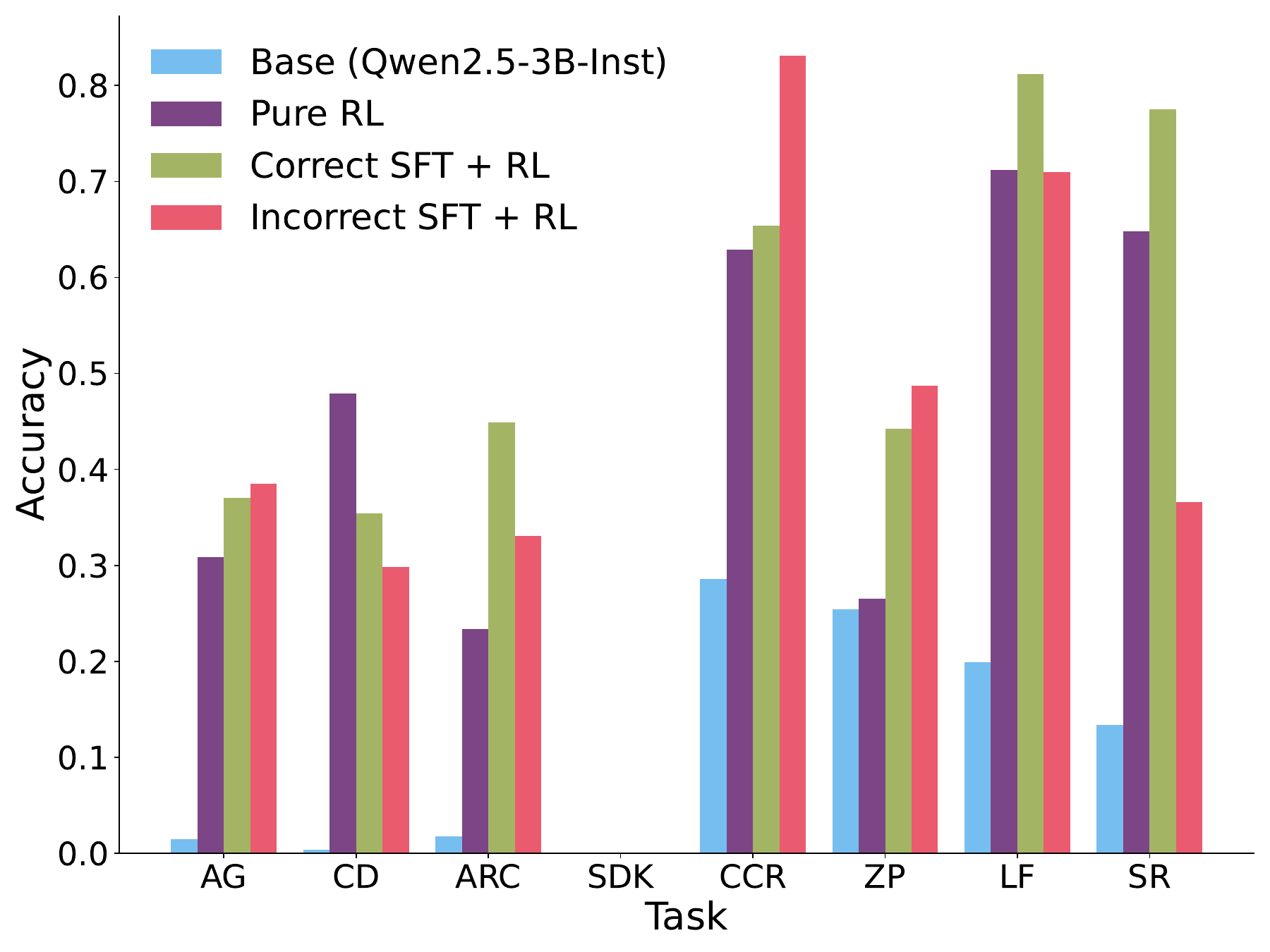}
    \caption{In-domain evaluation comparison across training methods.}
    \label{fig:short-cots-eval}
\end{subfigure}
\hfill
\begin{subfigure}[t]{0.48\columnwidth}
    \centering
    \includegraphics[
        width=\linewidth,
        trim=10cm 0cm 10cm 0cm,
        clip
    ]{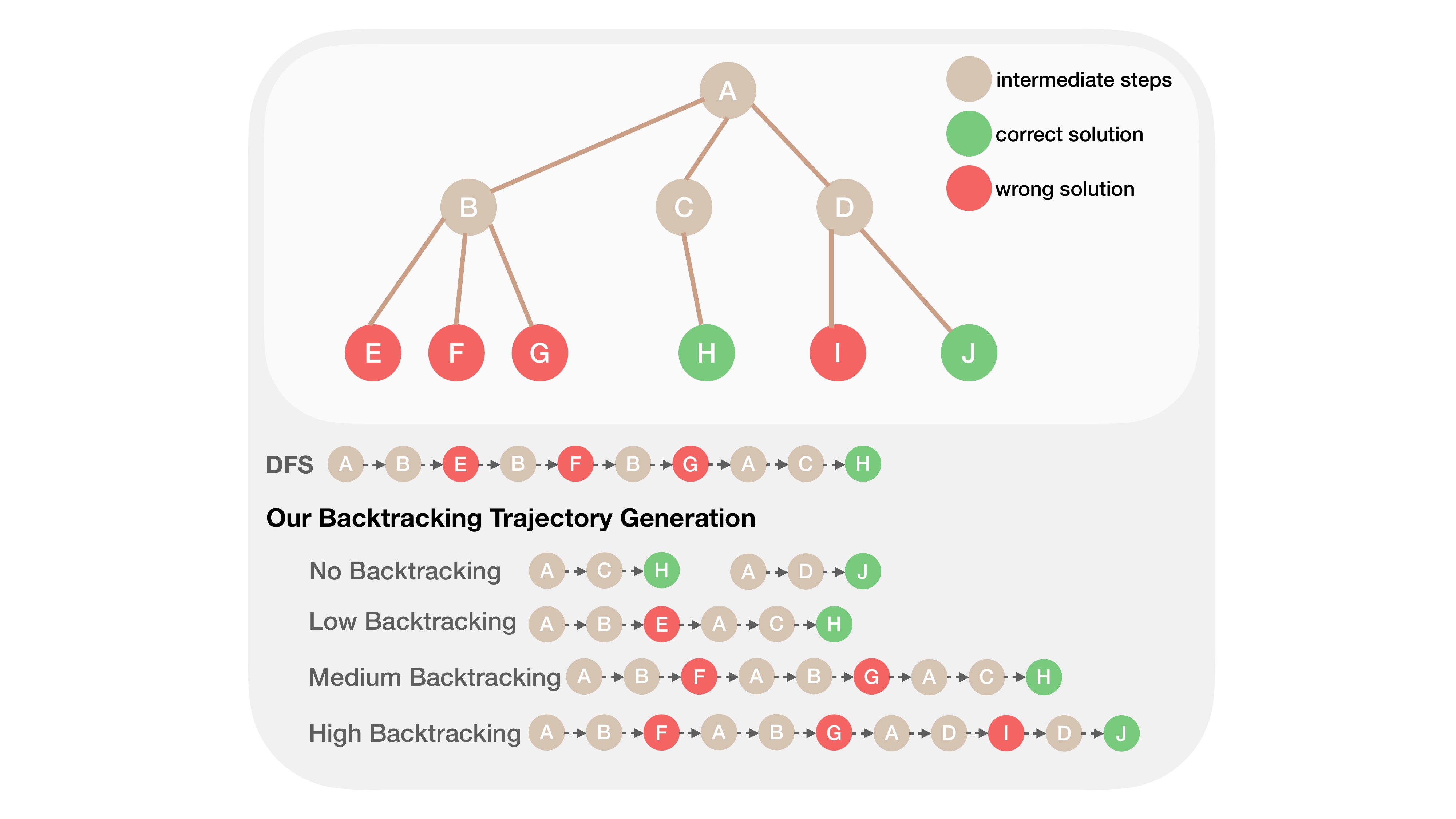}
    \caption{Synthetic backtracking data generation.}
    \label{fig:backtrack_tree}
\end{subfigure}

\caption{
\textbf{(Left)} In-domain evaluation comparison between base model, pure RL, correct self-sampled SFT + RL, and incorrect self-sampled SFT + RL. 
\textbf{(Right)} Backtracking trajectory generation with varying depths.
}
\label{fig:combined}
\end{figure}


In this series of experiments, we first perform specialized pure RL training starting from Qwen2.5-3B-Instruct \citep{qwen2.5} across 8 reasoning datasets \citep{stojanovski2025reasoning}. After obtaining 8 RL'ed models, we subsequently collect their reasoning trajectories on the same tasks they are finetuned on, but with problems generated from a different seed (different problems). We purposefully separate the trajectories that lead to correct answer and those that lead to incorrect answers, from which we further perform the complete SFT + RL post-training pipeline starting again from vanilla Qwen2.5-3B-Instruct, and obtain 15 specialized models (7 RL'ed models initialized from correct SFT, and the other 8 initialized from incorrect SFT). We maintain identical gradient steps between pure RL and self-sampled SFT + RL to ensure fair comparison. A comprehensive evaluation is subsequently performed for all 23 models (three models per task with the exception of Sudoku, which we discuss below) in Figure~\ref{fig:short-cots-eval}.

\paragraph{Self-sampled CoTs are effective warm-ups} Short self-sampled CoT warm-up consistently demonstrated performance gains across nearly all tasks compared to pure RL in Figure~\ref{fig:short-cots-eval}. Specifically, we observed significant performance improvements on tasks such as ARC and List Functions. It is also worth noting that even incorrect trajectories warm-up can improve downstream RL performance on some of the tasks. We will discuss more details in later section.

Notably, tasks like Sudoku still posed substantial challenges, highlighting the limitations of self-sampled CoTs for deeper, combinatorial reasoning. In fact, the cold-start RL pipeline effectively breaks down for Sudoku because the model cannot self-generate the necessary correct SFT initialization data, which illustrates a fundamental brittleness of the cold-start RL paradigm. These results underscore the incremental yet limited nature of cold-start RL, motivating further exploration of more sophisticated strategies.

\section{Pushing beyond boundaries with backtracking}
\label{sec:pushing-boundries-backtrack}


    Recognizing the limitations identified in previous methods, particularly the failure of both pure RL and short‑CoT warm‑ups on combinatorially hard tasks such as Sudoku, we propose a backtracking‑centric training paradigm to push models beyond their baseline reasoning limits. Empirical analyses of the strongest reasoning models reveal spontaneous use of backtracking (e.g., self-correct, multiple attempts), suggesting that explicit exposure to backtracking traces during SFT can seed deeper search policies that subsequent RL can refine. We therefore hypothesize that integrating controlled amounts of backtracking into the SFT stage will enable models to internalize more powerful search strategies and unlock substantial gains on challenging tasks.

\begin{figure}[t]
    \centering
    \includegraphics[width=\linewidth]{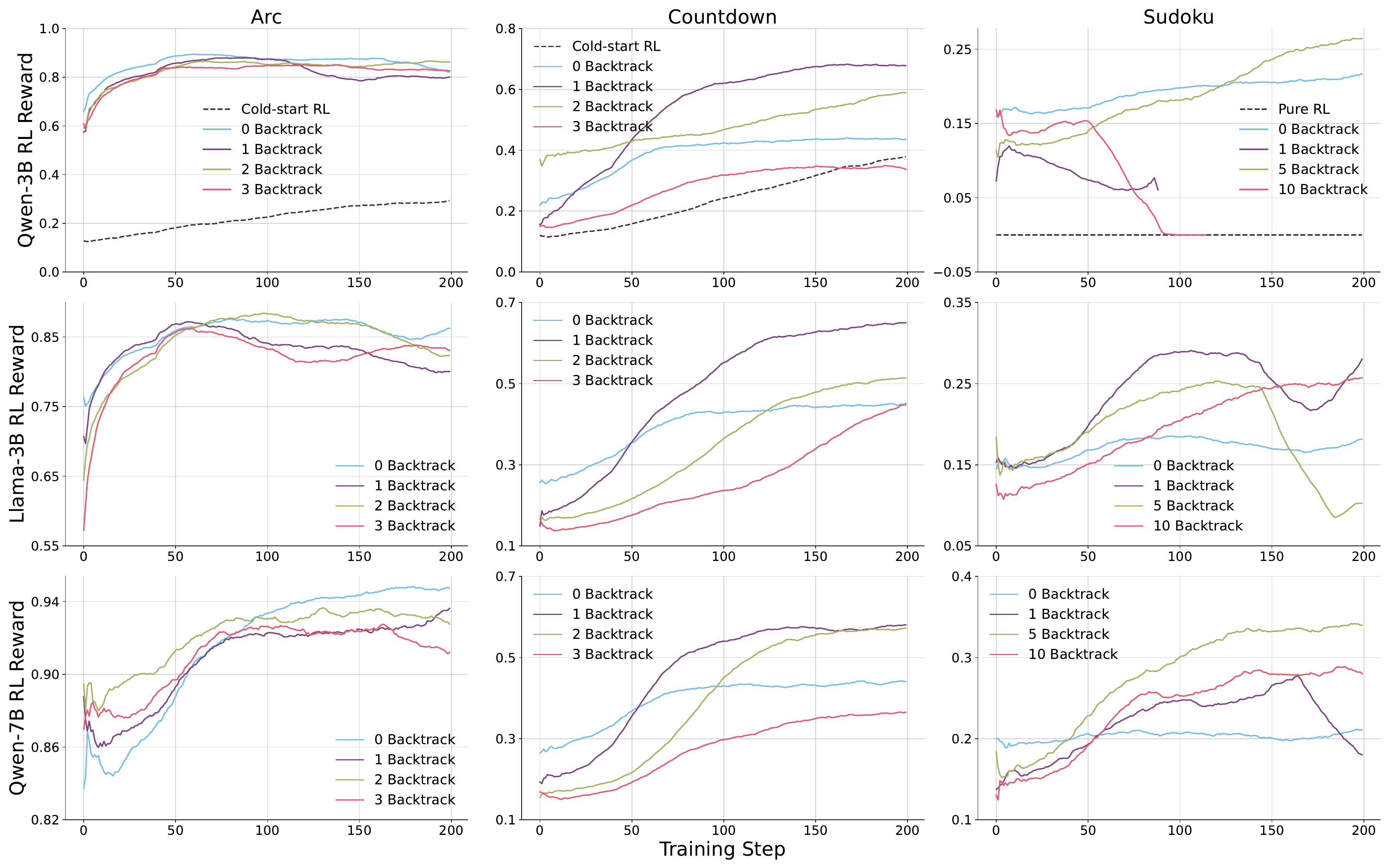}
    \caption{RL training rewards for Arc-1D, Countdown and Sudoku across Qwen2.5-3B-Instruct, Llama3.2-3B-Instruct, and Qwen2.5-7B-Instruct. We plot the RL reward curves corresponding to different backtracking warm-up strategies. The results indicate that distinct tasks require specific backtracking configurations for optimal performance.}
    \label{fig:syn-3-task-rewards}
\end{figure}

    \subsection{Building synthetic backtracking datasets} To systematically evaluate the impact of backtracking, we selected three representative reasoning tasks—Countdown, Arc 1D, and Sudoku—to construct synthetic datasets with varying numbers of backtrack, as well as optimal trajectories, and study whether, and how RL amplifies such behavior. The three tasks are selected because they are the more challenging subset based on vanilla model's baseline performance in Figure~\ref{fig:short-cots-eval}. 

To construct synthetic datasets for Countdown and Sudoku, we use DFS solver and create a tree structure where each node represents an intermediate step (e.g., an operation between two numbers for Countdown; fill one grid with a valid number for Sudoku).
Then we start from the solution node back to the root node to build optimal solutions, and include incorrect branches as detours and backtracks. For Arc 1D,  we design a heuristic search that applies hand-crafted transformation functions with step-wise, verbalized reasoning. A detour or backtrack is created by choosing an incorrect transformation and then retrying (examples in Appendix~\ref{appendix:backtrack-trajectories}).

In all three tasks, we define a $k$-backtrack trajectory as one containing $k$ explicit recovery events, where the reasoning enters an incorrect branch, returns to the most recent valid decision point, and then resumes along the correct solution path.

\begin{table*}[t]
  \caption{Evaluation accuracies ($\%$) for Qwen2.5-3B-Instruct, Qwen2.5-7B-Instruct, and Llama3.2-3B-Instruct models post-trained on synthetic datasets across zero backtracks, low backtrack (1 backtrack), medium backtrack (2 backtracks for Arc and Countdown, 5 for Sudoku), and high backtrack (3 backtracks for Arc and Countdown, 10 for Sudoku). QwQ-32B is included as a baseline.}
  \label{table:syn-backtrack-acc}
  \begin{center}
    \begin{small}
      \begin{sc}
        \setlength{\tabcolsep}{3pt}
        \resizebox{\textwidth}{!}{%
        \begin{tabular}{lcccccccccccc}
          \toprule
          & \multicolumn{4}{c}{\textbf{Arc-1D}} & \multicolumn{4}{c}{\textbf{Countdown}} & \multicolumn{4}{c}{\textbf{Sudoku}} \\
          \cmidrule(lr){2-5} \cmidrule(lr){6-9} \cmidrule(lr){10-13}
          Data & \shortstack{Qwen\\3B} & \shortstack{Qwen\\7B} & \shortstack{Llama\\3B} & \shortstack{QwQ\\32B}
               & \shortstack{Qwen\\3B} & \shortstack{Qwen\\7B} & \shortstack{Llama\\3B} & \shortstack{QwQ\\32B}
               & \shortstack{Qwen\\3B} & \shortstack{Qwen\\7B} & \shortstack{Llama\\3B} & \shortstack{QwQ\\32B} \\
          \midrule
          Zero & \textbf{90.1} & \textbf{94.1} & \textbf{91.5} & -
                  & 38.9 & 38.6 & 37.6 & -
                  & 14.4 & 11.8 & 9.6 & - \\
          Low     & 86.9 & 92.1 & 85.3 & -
                  & \textbf{69.7} & \textbf{59.4} & \textbf{63.0} & -
                  & 0.0 & 20.8 & 23.4 & - \\
          Medium  & 86.1 & 89.3 & 87.2 & -
                  & 60.8 & 57.7 & 48.2 & -
                  & \textbf{28.9} & \textbf{32.1} & 25.2 & - \\
          High    & 84.8 & 91.4 & 82.6 & -
                  & 32.4 & 31.4 & 44.4 & -
                  & 0.0 & 28.0 & \textbf{26.9} & - \\
          Baseline & - & - & - & 24.0
                   & - & - & - & 51.5
                   & - & - & - & 0.0 \\ 
          \bottomrule
        \end{tabular}
        }
      \end{sc}
    \end{small}
  \end{center}
\end{table*}

\subsection{Quantifying the Impact of Backtracking in SFT Warm-up} Our controlled experiments reveal that the exposure to backtrack systematically improves RL training efficiency as shown in Figure~\ref{fig:syn-3-task-rewards}. However, the number of backtracks needed depends on the difficulty of the problem:
\begin{enumerate}
    \item \textbf{Countdown — Moderate.} Every puzzle provides 4–6 numbers and a target value, producing a medium-sized arithmetic search tree. Although vanilla Qwen2.5-3B-Instruct can generate plausible attempts, it only solves 0.4\% of the problems. We found that using exactly \textbf{one} backtrack is the most optimal.
    \item \textbf{Arc 1D — Moderate/Easy.} Tasks are solved by applying a small pool of grid-transformation heuristics; the same baseline attains 1.8\%  accuracy. Using \textbf{zero} backtrack is optimal.
    \item \textbf{Sudoku — Hard.} Each board contains 30–60 empty cells; every cell admits multiple candidates, yielding an astronomically large search space. A vanilla Qwen2.5-3B-Instruct baseline solves none of the test instances. Using \textbf{five or more} backtracks is optimal.
\end{enumerate}

Our controlled experiments with both Qwen and Llama model families confirm a consistent pattern: harder combinatorial problems require deeper backtracking traces to seed RL, whereas easier tasks are best served by shallow or even optimal demonstrations.

For \textbf{Countdown}, RL initialized with one backtrack achieves the highest reward curve for all three evaluated models (Figure~\ref{fig:syn-3-task-rewards}), whereas zero backtracks or more frequent backtracking produce consistently worse training trajectories. This suggests that a one-backtrack initialization enables models to internalize and execute efficient search strategies. Evaluation results confirm this: the one-backtrack Qwen2.5-3B, Qwen2.5-7B, and Llama3.2-3B models attain $69.7\%$, $59.4\%$, and $63.0\%$ accuracy respectively, all outperforming QwQ-32B \citep{qwq32b} at $51.5\%$ with an 8k-token context window (Table~\ref{table:syn-backtrack-acc}). By contrast, their corresponding zero-backtrack models achieve only $38.9\%$, $38.6\%$, and $37.6\%$ respectively.

For \textbf{Sudoku}, initializing PPO with five or ten backtracks is necessary for stable, effective training, depending on the model and scale. Too few or too many backtracks can produce unstable or degraded training. For both Qwen models, five-backtrack training performs best on Sudoku: Qwen2.5-3B achieves $28.9\%$ versus $14.4\%$ with zero backtracks, while Qwen2.5-7B achieves $32.1\%$ versus $11.8\%$. For Llama, higher backtracking counts leads to substantially better evaluation performance, with the ten-backtrack model achieving markedly higher accuracy than the one and five-backtrack variant (Table~\ref{table:syn-backtrack-acc}). The best-performing backtracking configuration for each of the three models outperforms the larger and more powerful reasoning model QwQ-32B, which achieves $0.0\%$ (Table~\ref{table:syn-backtrack-acc}). Notably, pure RL fails to train on Sudoku, with reward trajectories stagnating at 0.1, highlighting the necessity of our method.

For \textbf{Arc 1D}, the easiest of the three tasks, performance is generally lower when backtracking is introduced. Zero-backtrack training performs best for all three models, achieving $90.1\%$, $94.1\%$, and $91.5\%$ accuracy for Qwen2.5-3B, Qwen2.5-7B, and Llama3.2-3B, respectively, substantially exceeding QwQ-32B’s $24.0\%$

See Appendix~\ref{appendix:additional-validation} for additional model and task scope ablations.

\section{Additional analysis of interplay between SFT and RL}
\label{sec:additional-analysis}
In this section, we document additional findings and include analysis throughout our goal of breaking the base boundary of reasoning.

\subsection{Incorrect trajectories are still useful}
\label{sec:rl-correctness}

Models initialized from incorrect reasoning CoTs display similar behaviors with those from correct trajectories. Figure~\ref{fig:short-cots-eval} shows that incorrect SFT models post RL see the same trend of performance increase or decrease with correct SFT models on in-domain evaluations, with the only exception being self reference task. While initially counter‑intuitive, this finding aligns with recent evidence that RL can quickly overwrite semantic errors provided the demonstrations respect the expected CoT format~\citep{setlur2024rl}. We conjecture that syntactic scaffolding alone primes the base model with an executable search blueprint that subsequent RL can refine. Such result also holds with the larger Qwen2.5-7B-Instruct on Countdown (Appendix~\ref{appendix:incorrect_sft_ablations}).

\subsection{RL is sensitive to shuffled SFT}
\begin{figure}[t]
  \centering

  \begin{subfigure}[t]{0.48\textwidth}
    \centering
    \includegraphics[width=\textwidth]{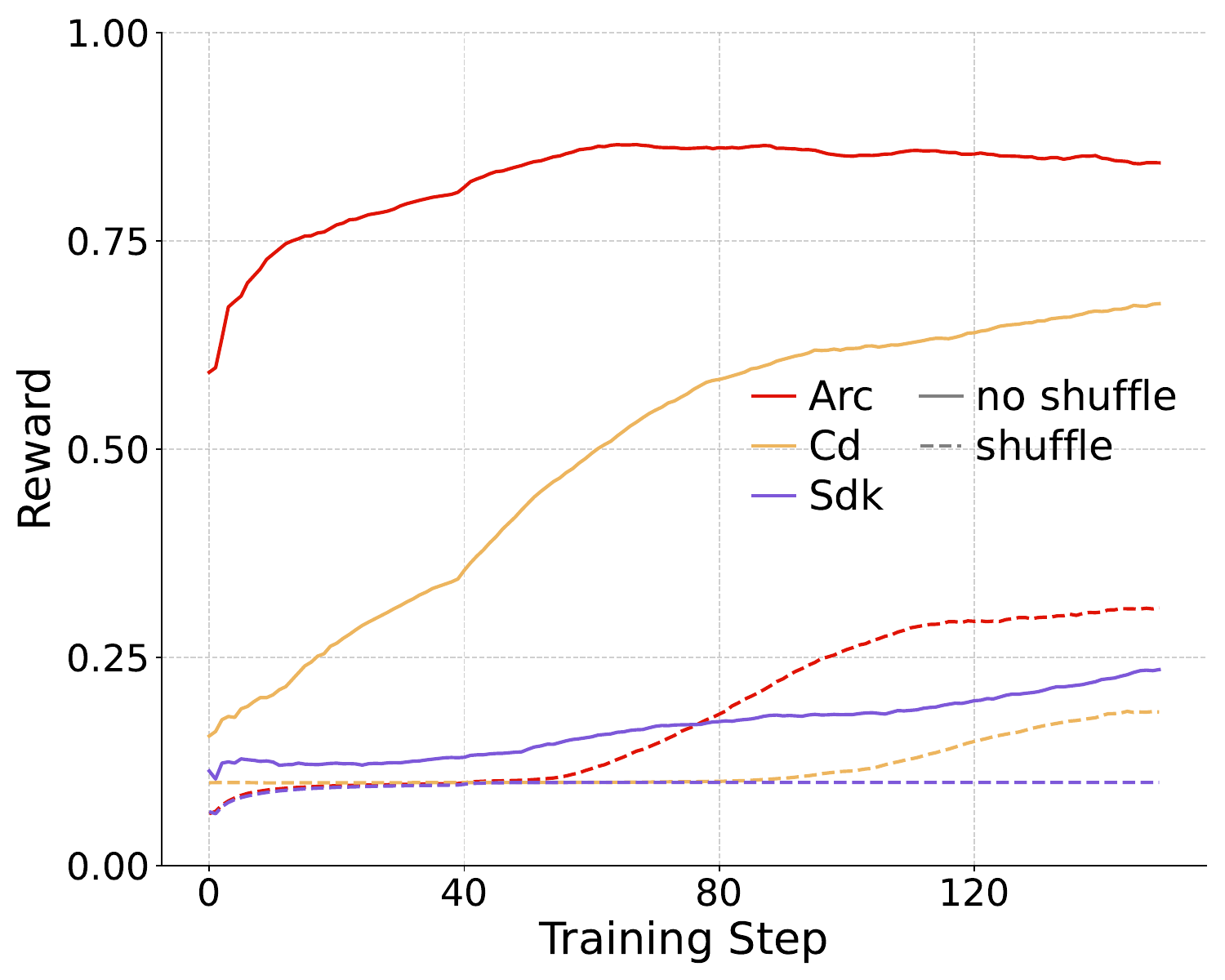}
    \caption{Reward curve (no shuffle vs shuffle)}
    \label{fig:all_shuffle}
  \end{subfigure}
  \hfill
  \begin{subfigure}[t]{0.48\textwidth}
    \centering
    \includegraphics[width=\textwidth]{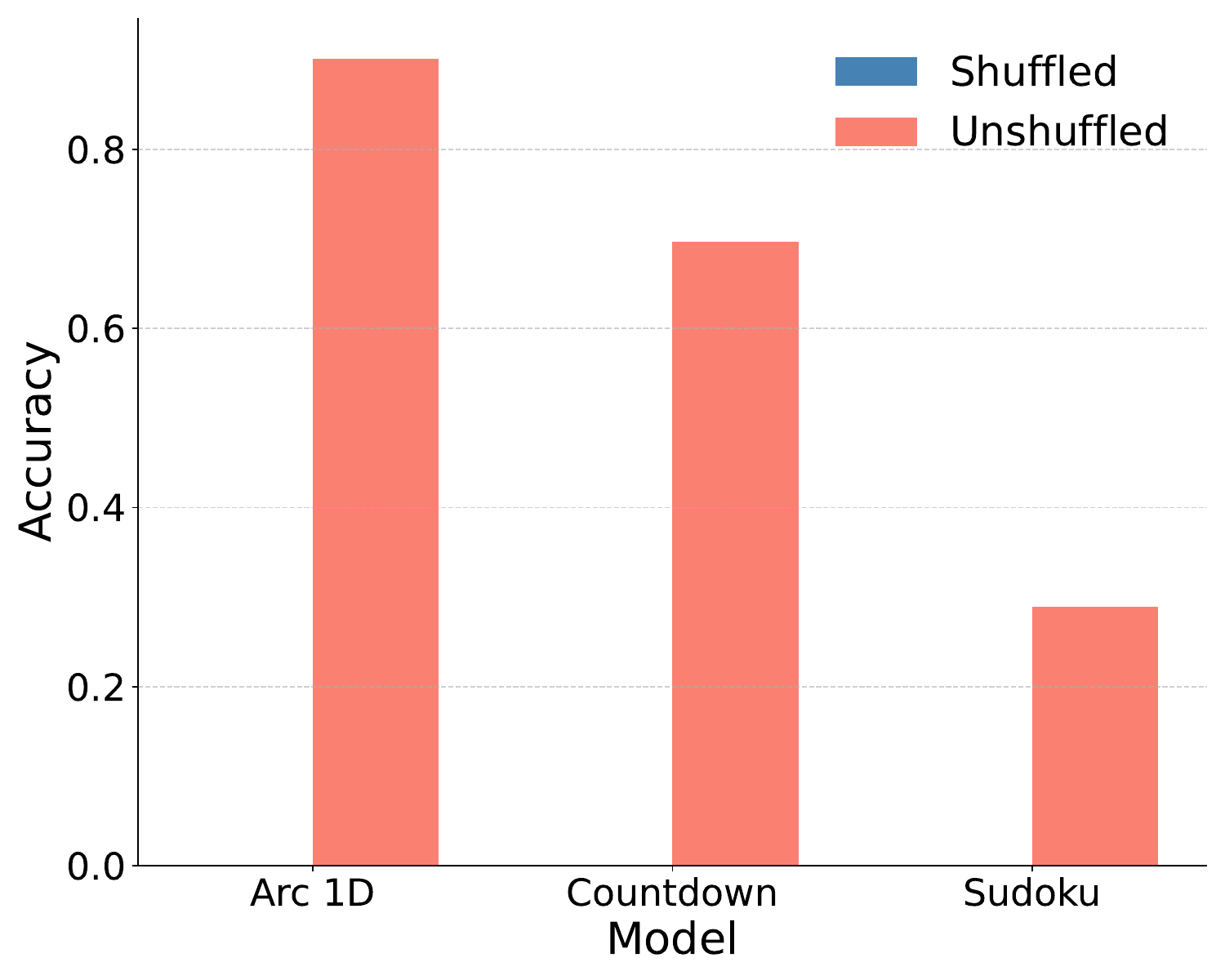}
    \caption{In domain evaluation accuracy}
    \label{fig:shuffle-rl}
  \end{subfigure}

  \caption{\textbf{(Left)} RL training reward trajectories for models post-trained on shuffled and non-shuffled synthetic datasets. \textbf{(Right)} In domain evaluation for each model post-trained on shuffled and not shuffled synthetic dataset of Arc 1D, Countdown, and Sudoku.}
\end{figure}

To probe the limits of semantic noise that an SFT warm‑up can tolerate, we deliberately broke the alignment between prompts and completions: within the best‑performing backtracking dataset in Section~\ref{sec:pushing-boundries-backtrack}. we randomly shuffled questions and their corresponding CoT‑answer pairs. This manipulation preserves syntactic structure—every example still contains well‑formatted trajectories while obliterating the underlying reasoning. From Figure~\ref{fig:all_shuffle}, we see that the training becomes extremely ineffective, suggesting that despite having the right reasoning structure, RL training is easily vulnerable to the internal inconsistency between prompt and completion during SFT. In fact, we find that, for the shuffled Countdown and Arc 1D runs in Figure~\ref{fig:all_shuffle}, models recover only a negligible reward signal after roughly 60–80 PPO updates before stagnating. These results contrast sharply with the resilience to incorrect but aligned CoTs observed in previous subsection, indicating that RL is highly sensitive to internal inconsistencies between prompts and a reasoning traces. Manual inspection of generated outputs (Appendix~\ref{appendix:model_outputs-shuffle}) reveals that the policy over‑fits to spurious prompt–completion pairings that PPO cannot unlearn.


\section{Related Work}
\subsection{Test time scaling}

    Recent work has shown scaling up language model output length during test time can be more effective than scaling pretraining \citep{snell2024scaling, wang2024mixture, jiang2023llm, saad2024archon}. As language models become increasingly larger and pre-train on increasingly large corpora; this third axis of scaling offers promising gains in performance while remaining budget friendly \citep{wang2025think}. There are mainly two directions for scaling up: parallel sampling and sequential search. Common parallel sampling methods such as best-of-n rely on LLMs to propose answers independently N times \citep{li2024common}; whereas sequential search generates each response in sequence conditioned on previous attempt \citep{qu2024recursive, feng2023alphazero, gandhi2024stream, yang2025step}. There have also been attempts to combine the two and incorporate external verifiers through algorithms like Monte-Carlo Tree Search \citep{guan2025rstar, feng2023alphazero}. Our work extends this line of research by examining the optimal training mixtures for efficiently scaling up model generation for reasoning.

\subsection{Reinforcement Learning for reasoning}

     A novel dimension of scaling inference-time compute for language models has recently gained prominence through the emergence of specialized "reasoning models" trained via reinforcement learning (RL). Prior works predominantly explored on-policy and off-policy RL methods \citep{zelikman2022star, zelikman2024quiet, kazemnejad2025vinepporefiningcreditassignment, phan2023training} and saw moderate successes. However, recent approaches such as DeepSeek-R1 \citep{guo2025deepseek} have revitalized interest in RL training utilizing verifiable rewards (RLVR), driven by Proximal Policy Optimization (PPO) and its memory-efficient variant, Group Relative Policy Optimization (GRPO) \citep{shao2024deepseekmath}. A key driver behind RLVR is the emphasis on rule-based rewards derived directly from final outputs, replacing traditional reward models. This new RL training paradigm has notably revealed critical "aha moments," where models spontaneously generate distinctive tokens such as "wait," indicative of active internalization of search strategies and optimized trajectory discovery. 
     
     One prevailing hypothesis is that such fine-tuning allows language models to internalize search algorithms during inference-time explicitly. Methods such as Stream of Search \citep{gandhi2024stream}, for instance, fine-tune models using linearized search trajectories, empowering continuous and coherent search capability within single-output generation \citep{yang2025step, lehnert2024beyond}. Alternatively, another hypothesis suggests that RL training utilizing verifiable rewards does not fundamentally induce novel reasoning skills but rather exploits and amplifies capabilities already learned during the pretraining phase \citep{yue2025does}. This claim is supported by observations that base pretrained models can achieve comparable or superior performance in metrics such as pass@k compared to their RL-trained counterparts. Nevertheless, the precise mechanisms behind these emergent reasoning capabilities remain poorly understood. Our research addresses this gap by systematically investigating how RL leverages SFT and the associated data mixture to unlock the reasoning capacities. 

     There are a few concurrent efforts investigating how fine-grained behavioral signals influence RL-trained reasoning models. \citet{gandhi2025cognitive} analyzes four behavioral cues, demonstrating that such cues can accelerate RL convergence. We perform controlled experiments on more tasks and study various factors of SFT mixtures. \citet{qin2025backtrack} examines whether explicit backtracking traces improve performance across CountDown and Sudoku. We go further by rigorously controlling trace correctness, format, and synthetic backtrack frequency. Finally, \citet{zhao2025echo} explores how variations in pre‑training corpus composition interact with subsequent post-training, whereas we emphasize on how post-training affects the reasoning ability.

\section{Conclusion}

    In this work, we thoroughly inspect the interplay between SFT and RL for reasoning tasks through controlled experiments, emphasizing the role of training-data composition towards effective RL post-training. We found that Qwen2.5-3B-Instruct models, when performing pure RL, demonstrates two distinct reasoning pattern: verbalized searching and backtracking that iteratively tries to find the correct answer, and concise thinking which solves the problem in one go. To solve combinatorially hard problems like Sudoku, we incorporate synthetic backtracking into our training paradigm and evaluate it across Qwen2.5-3B-Instruct, Qwen2.5-7B-Instruct, and Llama3.2-3B-Instruct. Our results show that the optimal backtrack depth increases with task difficulty, yielding up to a $32.1\%$ accuracy improvement over the base model on Sudoku. Further, we have empirical evidence demonstrating that, when initializing RL training from models that have seen incorrect demonstrations, training can still be stable and effective. Leveraging the characteristics of pretrain and SFT data to continually scale model performance through RL remains an intriguing direction for future works.

\section*{Acknowledgments}
Junlin Wang is partially supported by through Grant G-23- 2137070 from the Learning Engineering Virtual Institute to the University of Florida and its partner institutions, including Duke.

\bibliography{colm2026_conference}
\bibliographystyle{colm2026_conference}

\newpage

\appendix

\section{Usage of LLM}
We utilize LLM to assist some of the paper, specifically polish grammar and sentences. We also utilized LLMs for simple code development, for example writing a initial evaluation code. Lastly, we also use LLM to write the code for figure generation.

\section{Reasoning Games}
\label{appendix:reason-games}
\paragraph{Countdown}
The goal is to construct an arithmetic expression using a set of numbers to reach a specific target value. Each number can be used at most once, and operations are limited to basic arithmetic: addition, subtraction, multiplication, and division. 

\paragraph{Sudoku}
The objective is to fill a 9$\times$9 grid so that each row, column, and 3$\times$3 subgrid contains all the digits from 1 to 9 exactly once. 

\paragraph{Arc 1D}
The objective is to learn and apply a transformation rule that maps a one-dimensional input grid to a corresponding output grid, given several input–output examples.

\paragraph{Advanced Geometry}
Advanced Geometry contains three sub-tasks. In the angle measurement sub-task, the goal is to compute the internal angle at a specific vertex of a triangle, given the input of three vertex coordinates in the Cartesian plane. In the orthocenter sub-task, the goal is to determine the orthocenter of a triangle—the point where all three altitudes intersect, given the input of the triangle’s three vertex coordinates. In the incircle radius sub-task, the goal is to compute the radius of the incircle of a triangle, given the input of the coordinates of its three vertices.

\paragraph{Color Cube Rotation}
In the game of Color Cube Rotation, the goal is to determine the color of a specific side of a cube after a sequence of 3D rotations, given the input of its initial face-color configuration.

\paragraph{List Functions}
The goal is to generate an output list by identifying and applying an implicit rule that maps each input list to its corresponding output, given examples.

\paragraph{Zebra Puzzles}
The goal is to determine a specific attribute—such as the name of the person in a given house—based on a set of logical constraints, given the input of several people and their distinct characteristics (e.g., favorite drink, pet, or phone).

\paragraph{Self Reference}
 The goal is to determine how many consistent truth assignments exist for a set of self-referential statements, given the input of seven logically interconnected claims. Each statement makes assertions about the truth or falsity of other statements (or the total number of true/false ones).

\section{Model training details}
\label{appendix:rl-details}
For RL, we investigate both PPO and GRPO. We use learning rate of $1e-6$ for both actor and critic, KL control coefficient of 0 (no KL penalty), gradient clipping of 1.0, entropy coefficient of 0.001, value coefficient of 1.0, gae lambda of 1.0, gamma of 1.0. These hyperparameters are held constant across all runs, with the only difference being the response length to accommodate for problems with longer solution trajectories such as Sudoku. All experiments are done on 4 A100s.

\section{Reward Setup}
\label{appendix:reward-setup}
\paragraph{Countdown} We expect solution to not include the equal sign
\paragraph{Advanced Geometry} For angle measure task: we expect a solution string rounded to two decimal places with degrees symbols attached. Failure to include degree symbol, or not rounding to two decimal places will be deemed incorrect. For orthocenter task: we expect a solution to give the coordinate either in the form “(1.000 2.000)” or “(1.000, 2.000)” or “(1.000,2.000)”. Failure to include “()” or not rounding to 3 decimal places will be deemed incorrect. For incircular radius task: we expect a solution string in the format of “1.123”. Failure to round to 3 decimal places will be deemed incorrect.
\paragraph{Arc 1D} We expect a solution string in the format of “1 2 3”, other formats such as “1, 2, 3” will be deemed incorrect.
\paragraph{Sudoku} We expect a solution string in the format of “1 2 3\textbackslash n4 5 6\textbackslash n7 8 9”, other formats such as “1, 2, 3\textbackslash n4, 5, 6\textbackslash n7, 8, 9\textbackslash n” will be deemed incorrect and only receive format score. Parsing includes generating a list of lists from the solution string.
\paragraph{Color Cube Rotation} The solution parser is case insensitive. We expects a solution string in the format of “cyan” or “Cyan”. Parsing includes lower case all letter.
\paragraph{Zebra Puzzles} We expect a solution string in the format of “bob” or “Bob”. Parsing includes lower case all letters.
\paragraph{List Functions} We expect a solution string either in the format of “[1 2 3]” or “[1, 2, 3]” or “[1,2,3]”. Failure to include “[]” will be deemed incorrect. Will parse to a numpy array.
\paragraph{Self Reference} We expect a solution in the format of a single digit like "4". Parsing includes converting the solution string to an integer.

\section{Model and Task Scope Ablations}
\label{appendix:additional-validation}

\subsection{Robustness to a newer model generation}
\label{appendix:newer-model}

The main experiments show a consistent pattern across Qwen2.5-3B-Instruct, Qwen2.5-7B-Instruct, and Llama3.2-3B-Instruct: zero backtracks perform best on Arc-1D, one backtrack performs best on Countdown, and substantially deeper backtracking performs best on Sudoku, although the exact Sudoku depth varies by model. To test whether the Countdown result also holds in a newer model generation, we repeat the same backtracking experiment in Section~\ref{sec:pushing-boundries-backtrack} with Qwen3.5-2B \citep{qwen3.5}.

As shown in Table~\ref{tab:qwen35-countdown}, the one-backtrack warm-up achieves the highest pass@1 among the tested depths: $47.6\%$, compared with $35.2\%$ for zero backtracks, $32.1\%$ for two backtracks, and $29.3\%$ for three backtracks.

\begin{table}[ht]
  \caption{Countdown pass@1 accuracy after synthetic-backtracking SFT warm-up and RL post-training with Qwen3.5-2B.}
  \label{tab:qwen35-countdown}
  \begin{center}
    \begin{tabular}{lcccc}
      \toprule
      Model & 0 Backtrack & 1 Backtrack & 2 Backtrack & 3 BT \\
      \midrule
      Qwen3.5-2B & 35.2 & \textbf{47.6} & 32.1 & 29.3 \\
      \bottomrule
    \end{tabular}
  \end{center}
\end{table}

\subsection{Preliminary generalization to MATH}
\label{appendix:math-backtracking}

Our synthetic backtracking datasets for Countdown and Sudoku are generated from symbolic solvers, while Arc-1D uses a heuristic search. Constructing an exhaustive search tree is considerably harder for open-ended domains such as math reasoning or coding. However, our method emphasizes on the metacognitive behavior that backtracking captures: recognizing that the current reasoning state is flawed, returning to a prior valid state, and continuing from there.

To test whether the backtracking-centric recipe generalizes beyond puzzle-style tasks, we conduct a preliminary experiment on MATH \citep{hendrycks2021measuring}. We use the provided step-by-step gold solutions as zero-backtrack trajectories. To construct the one-backtrack trajectories, we prompt a stronger Qwen3.6-35B-A3B \citep{qwen36_35b_a3b} teacher model to rewrite them into one-backtrack trajectories by injecting an intermediate error, explicit recognition, and recovery to the valid path. Therefore, we constructed our proxy search tree trajectories via distillation from the teacher model. After post-training, one-backtrack warm-up improves pass@1 over the zero-backtrack warm-up from $20.78\%$ to $25.74\%$, providing initial evidence that backtracking-style warm-up can extend to broader tasks where solver based search trees are not available.

\begin{table}[ht]
  \caption{MATH pass@1 accuracy after post-training Qwen2.5-0.5B-Instruct with zero- and one-backtrack warm-ups.}
  \label{tab:math-backtracking}
  \begin{center}
    \begin{tabular}{lcc}
      \toprule
      Task / Model & 0 Backtrack & 1 Backtrack \\
      \midrule
      MATH / Qwen2.5-0.5B-Instruct & 20.78 & \textbf{25.74} \\
      \bottomrule
    \end{tabular}
  \end{center}
\end{table}

\section{Incorrect vs. Correct Warmup Comparison}
\label{appendix:incorrect_sft_ablations}

\begin{table}[ht]
  \centering
  \caption{
    Countdown accuracy after RL with correct and incorrect self-sampled SFT warm-ups for Qwen2.5-7B-Instruct.
  }
  \label{tab:self-sampled-scaling}
  \begin{small}
    \begin{tabular}{lllc}
      \toprule
      Model & Task & SFT trajectory & Pass@1 (\%) \\
      \midrule
      Qwen2.5-7B-Instruct   & Countdown & Correct   & 37.5 \\
      Qwen2.5-7B-Instruct   & Countdown & Incorrect & \textbf{40.5} \\
      \bottomrule
    \end{tabular}
  \end{small}
\end{table}

\section{Example Model Outputs}
\label{appendix:model_outputs}
\subsection{List Functions Pure RL model outputs}
\label{appendix:model_output-lf-short}
\begin{PromptBox}[title=Pure RL Outputs: List Functions]
\begin{verbatim}
<|im_start|>assistant
Let me solve this step by step.
<think> The rule in the examples seems to be reversing the order 
of the elements in the list. Let's apply this rule to the input 
list [74, 72, 49, 43, 41, 3, 30, 85, 48]. Reversing [74, 72, 49, 
43, 41, 3, 30, 85, 48] gives us [48, 85, 30, 3, 41, 43, 49, 72, 
74]. </think>
<answer> [48, 85, 30, 3, 41, 43, 49, 72, 74] </answer><|im_end|>
\end{verbatim}
\end{PromptBox}

\subsection{Countdown Pure RL model outputs}
\label{appendix:model_output-cd-short}
\begin{PromptBox}[title=Pure RL Outputs: Countdown]

\begin{verbatim}
<|im_start|>assistant
Let me solve this step by step.
<think> First, I need to create an expression that equals 194 
using the numbers 14, 17, 2, 45, and 8. Let's try some 
combinations. If I add 45 and 17, I get 62. Then if I add 14 to 
62, I get 76. Adding 8 to 76 gives me 84. If I add 2 to 84, I 
get 86. I need to reach 194, so I need to try different 
combinations. Let's try 45 * 2 = 90, then 90 + 17 = 107, then 
107 + 14 = 121, then 121 + 8 = 129. This doesn't work. I need 
to find a better combination. Let's try 17 * 8 = 136, then 136 
+ 45 = 181, then 181 + 14 = 195, then 195 - 2 = 193. Still not 
194. I need to find a different approach. Let's try 2 * 14 = 
28, then 28 + 45 = 73, then 73 + 17 = 90, then 90 + 8 = 98. 
Still not 194. I need to find a different combination. Let's 
try 45 * 2 = 90, then 90 + 14 = 104, then 104 + 17 = 121, then 
121 + 8 = 129. Still not 194. I need to find a different 
approach. Let's try 17 * 14 = 238, then 238 - 45 = 193, 
then 193 + 8 = 201, then 201 - 2 = 199. Still not 194. I need 
to find a different combination. Let's try 14 * 2 = 28, then 
28 + 45 = 73, then 73 + 17 = 90, then 90 + 8 = 98. Still not 
194. I need to find a different approach. </think>
\end{verbatim}
\end{PromptBox}

\subsection{Countdown RL model from shuffled synthetic data output}
\label{appendix:model_outputs-shuffle}
\begin{PromptBox}[title=Shuffled RL Outputs: Countdown]
\begin{verbatim}
<|im_start|>assistant
Let me solve this step by step.
<think>
Step 1: 51 - 23 = 28. Step 2: 28 * 36 = 1008. Step 3: 1008 - 
57 = 951. Step 4: 951 - 48 = 885. Wait, this doesn't lead to 
the correct solution. 885 is not the correct answer. Let me 
go back to step 2 and keep thinking from there.Step 3: 1008 
- 27 = 981. Step 4: 981 - 48 = 933. This matches the problem 
statement. This is the solution.
</think>

<answer>51 - 36 * 36 - 57 - 48 - 27</answer><|im_end|>
\end{verbatim}
\end{PromptBox}

\section{Example Backtracking Trajectories}
\label{appendix:backtrack-trajectories}
\begin{PromptBox}[title=Countdown]
\begin{verbatim}
Step 1: 47 - 27 = 20. Step 2: 20 * 10 = 200. Step 3: 200 - 47 = 153. Wait, this 
doesn't lead to the correct solution. 153 is not the correct answer. Let me 
go back to step 2 and keep thinking from there. Step 3: 200 - 76 = 124. 
This matches the problem statement. This is the solution.

<answer>10 * (47 - 27) - 76</answer>
\end{verbatim}
\end{PromptBox}

\begin{PromptBox}[title=Truncated Example of Sudoku]
\begin{verbatim}
Looking at the cell at (1,1). Based on the current board, row 1 is missing 
[3, 4, 6, 7, 8], column 1 is missing [1, 3, 4, 5, 6, 7, 9], and the 3x3 box is 
[1, 3, 4, 6, 7]. Taking the intersection of these sets, the possible values 
for cell (1,1) are [3, 4, 6, 7]. Let's try to fill cell (1,1) with 3.
Looking at the cell at (1,2). Based on the current board, row 1 is missing 
[4, 6, 7, 8], column 2 is missing [2, 3, 4, 6, 7, 8, 9], and the 3x3 box is 
[1, 4, 6, 7]. Taking the intersection of these sets, the possible values 
for cell (1,2) are [4, 6, 7]. Let's try to fill cell (1,2) with 4.
Looking at the cell at (1,3). Based on the current board, row 1 is missing 
[6, 7, 8], column 3 is missing [5, 6], and the 3x3 box is [1, 6, 7]. 
Taking the intersection of these sets, the possible values for cell 
(1,3) are [6]. Let's try to fill cell (1,3) with 6.
Looking at the cell at (1,6). Based on the current board, row 1 is missing 
[7, 8], column 6 is missing [1, 5, 7, 8], and the 3x3 box is [3, 5, 7, 8]. 
Taking the intersection of these sets, the possible values for cell (1,6) 
are [7, 8]. Let's try to fill cell (1,6) with 7.
Looking at the cell at (1,8). Based on the current board, row 1 is missing 
[8], column 8 is missing [1, 3, 4, 5, 7, 8], and the 3x3 box is 
[1, 2, 3, 4, 6, 7, 8]. Taking the intersection of these sets, 
the possible values for cell (1,8) are [8]. Let's try to fill cell (1,8) with 8.
Looking at the cell at (2,4). Based on the current board, row 2 is missing 
[1, 3, 4, 7], column 4 is missing [3, 4, 5, 6, 8, 9], and the 3x3 box is 
[3, 5, 8]. Taking the intersection of these sets, the possible values for 
cell (2,4) are [3]. Let's try to fill cell (2,4) with 3.
Looking at the cell at (2,7). Based on the current board, row 2 is missing 
[1, 4, 7], column 7 is missing [1, 2, 6, 8], and the 3x3 box is [1, 2, 3, 4, 6, 7]. 
Taking the intersection of these sets, the possible values for cell (2,7) are 
[1]. Let's try to fill cell (2,7) with 1.
Looking at the cell at (2,8). Based on the current board, row 2 is missing 
[4, 7], column 8 is missing [1, 3, 4, 5, 7], and the 3x3 box is [2, 3, 4, 6, 7]. 
Taking the intersection of these sets, the possible values for cell (2,8) are 
[4, 7]. Let's try to fill cell (2,8) with 4.
Looking at the cell at (2,9). Based on the current board, row 2 is missing 
[7], column 9 is missing [1, 3, 4, 6, 7, 8], and the 3x3 box is [2, 3, 6, 7]. 
Taking the intersection of these sets, the possible values for cell (2,9) are 
[7]. Let's try to fill cell (2,9) with 7.
Looking at the cell at (3,1). Based on the current board, row 3 is missing 
[1, 2, 3, 5, 6, 7, 8], column 1 is missing [1, 4, 5, 6, 7, 9], and the 3x3 box is 
[1, 7]. 
Taking the intersection of these sets, the possible values for cell (3,1) are 
[1, 7]. Let's try to fill cell (3,1) with 1. The current board is:
3 4 6 1 2 7 5 8 9
2 5 8 3 6 9 1 4 7
1 _ 9 _ _ 4 _ _ _
_ _ _ _ 7 3 _ 9 _
_ _ 3 2 8 _ 7 _ _
_ _ 1 _ _ _ _ 2 _
_ 1 2 _ _ 6 9 _ _
8 _ 4 7 9 2 3 _ 5
_ _ 7 _ _ _ 4 6 2
\end{verbatim}
\end{PromptBox}

\begin{PromptBox}[title=Arc-1d]
\begin{verbatim}
Looking at these three examples. Let’s identify all the blocks of sequences 
of consecutive 2s separated by gaps. Then let's try to recolor the left half 
of each block to 8, and try to keep the right half unchanged. 
Applying this rule to the test input, we get the output grid:
Example 1: 0 0 0 0 0 0 0 3 0 3 3 0
Example 2: 2 0 2 0 2 0 8 8 2 2 2 0
Example 3: 2 0 2 0 8 8 8 8 2 2 2 2
The outputs do not match the given example outputs. 
Let's try to find another pattern.

Let’s first identify the main block — a group of consecutive cells with the 
same color. It looks like the block is surrounded by noise: individual 
pixels of the same color placed away from the block. Let's try remove all 
the pixels by setting them to 0. Applying this rule to the test input, 
we get the output grid:
Example 1: 0 0 0 0 0 0 0 0 0 3 3 0
Example 2: 0 0 0 0 0 0 2 2 2 2 2 0
Example 3: 0 0 0 0 2 2 2 2 2 2 2 2
The outputs match the given example output. Let's apply this rule to 
the test input and obtain the answer.

<answer>0 8 8 8 8 0 0 0 0 0 0 0</answer>
\end{verbatim}
\end{PromptBox}

\section{Qwen2.5-3B-Instruct Training Token Counts}
\begin{table}[H]
\caption{Token counts for different datasets and training methods on Countdown.}
\label{tab:countdown_tokens}
\begin{center}
\begin{tabular}{lc}
\multicolumn{1}{c}{\bf Dataset / Method} & \multicolumn{1}{c}{\bf \# Output tokens} \\
\hline \\
Optimal synthetic dataset & 108.81 \\
1 backtrack dataset & 796.16 \\
2 backtrack dataset & 883.26 \\
3 backtrack dataset & 745.31 \\
Correct cold start RL & 625.03 \\
Incorrect cold start RL & 221.37 \\
Pure RL & 442.40 \\
\end{tabular}
\end{center}
\end{table}

\begin{table}[H]
\caption{Token counts for different datasets and training methods on Arc 1D.}
\label{tab:arc1d_tokens}
\begin{center}
\begin{tabular}{lc}
\multicolumn{1}{c}{\bf Dataset / Method} & \multicolumn{1}{c}{\bf \# Output tokens} \\
\hline \\
Optimal synthetic dataset & 270.43 \\
1 backtrack dataset & 473.64 \\
2 backtrack dataset & 664.81 \\
3 backtrack dataset & 889.47 \\
Correct cold start RL & 127.70 \\
Incorrect cold start RL & 352.76 \\
Pure RL & 118.80 \\
\end{tabular}
\end{center}
\end{table}

\begin{table}[H]
\caption{Token counts for different datasets and training methods on Sudoku.}
\label{tab:sudoku_tokens}
\begin{center}
\begin{tabular}{lc}
\multicolumn{1}{c}{\bf Dataset / Method} & \multicolumn{1}{c}{\bf \# Output tokens} \\
\hline \\
Optimal synthetic dataset & 4702.90 \\
1 backtrack dataset & 7028.17 \\
5 backtrack dataset & 7690.53 \\
10 backtrack dataset & 7607.92 \\
Correct cold start RL &  \\
Incorrect cold start RL &  \\
Pure RL & 449.55 \\
\end{tabular}
\end{center}
\end{table}


\section{Self Sampled CoTs model evaluations}
\label{appendix:full-cold-start-eval}

    \begin{table}[H]
        \caption{Evaluation accuracy (in percentage) of the Qwen2.5-Instruct and QwQ-32B baselines and the pure RL'ed Qwen2.5-3B models.
        The row names are the model names (or the task on which they were RL'ed), and the column names represent the evaluation tasks.}
        \label{tab:baseline_eval}
        \begin{center}
        \begin{tabular}{lcccccccc}
            \multicolumn{1}{c}{\bf Model} &\multicolumn{1}{c}{\bf AG} &\multicolumn{1}{c}{\bf CD} &\multicolumn{1}{c}{\bf ARC} &\multicolumn{1}{c}{\bf SDK} &\multicolumn{1}{c}{\bf CCR} &\multicolumn{1}{c}{\bf ZP} &\multicolumn{1}{c}{\bf LF} &\multicolumn{1}{c}{\bf SR} \\
            \hline \\
            Qwen2.5-3B-Instruct &  1.5 &  0.4 &  1.8 & 0.0 & 28.6 & 25.4 & 19.9 & 13.4 \\
            Qwen2.5-7B-Instruct &  5.2 &  1.9 &  6.4 & 0.0 & 28.1 & 38.8 & 31.4 & 13.8 \\
            QwQ-32B             & 34.4 & 51.5 & 24.0 & 0.0 & 13.5 & ---  & 74.8 & ---  \\
            AdvGeom             & 30.9 &  1.9 &  2.1 & 0.0 & 24.4 & 31.9 & 24.1 & 13.1 \\
            Countdown           &  4.3 & 47.9 &  3.3 & 0.0 & 22.1 & 14.3 & 21.4 &  8.6 \\
            Arc1D               &  0.1 &  1.0 & 23.4 & 0.0 & 24.1 & 31.9 & 29.9 & 15.6 \\
            Sudoku              &  1.2 &  0.2 &  2.5 & 0.0 & 25.9 & 26.9 & 21.7 & 14.8 \\
            ColorCube           & 14.9 &  1.9 &  3.0 & 0.0 & 62.9 & 33.1 & 24.7 & 10.5 \\
            Zebra               &  7.6 &  1.9 &  2.4 & 0.0 & 34.6 & 26.5 & 24.0 & 11.2 \\
            ListFunc            & 12.0 &  1.5 &  2.8 & 0.0 & 34.0 & 30.8 & 71.2 & 12.5 \\
            SelfRef             &  0.0 &  0.7 &  5.4 & 0.0 & 26.8 & 35.2 & 18.5 & 64.8 \\
        \end{tabular}
        \end{center}
    \end{table}

\begin{figure}[H]
  \centering
  \includegraphics[width=1\textwidth]{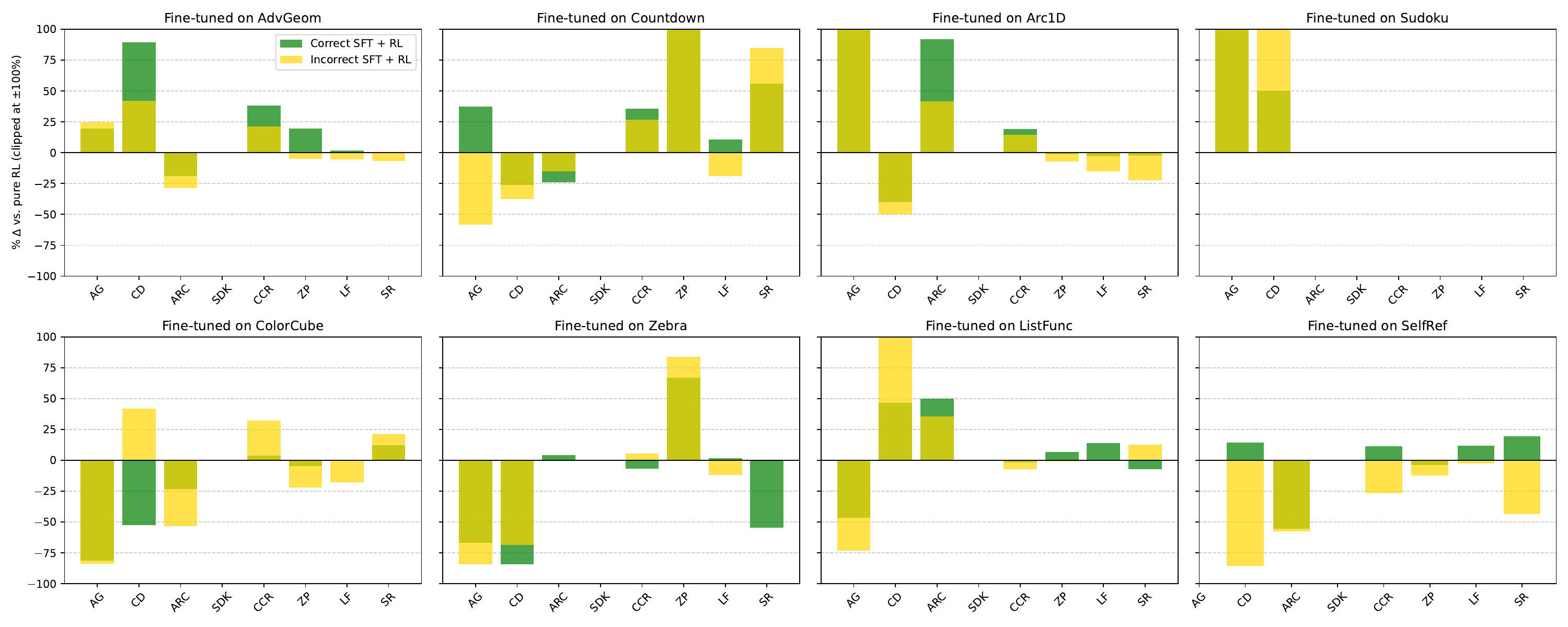}
  \caption{Correct and incorrect short self-sampled CoTs RL model evaluation. These plots measure the performance delta ($\Delta$) against a pure RL baseline. The results demonstrate that while some tasks (e.g., Countdown) promote broad reasoning generalization across the suite, others (e.g., ColorCube) exhibit narrow task-specific improvements or negative transfer to unrelated domains.}
  \label{appendix:more-delta-comparison}
\end{figure}

\section{Response Length Comparison}
\label{appendix:response-length-compare}
\begin{figure}[H]
  \centering
  \includegraphics[width=0.6\textwidth]{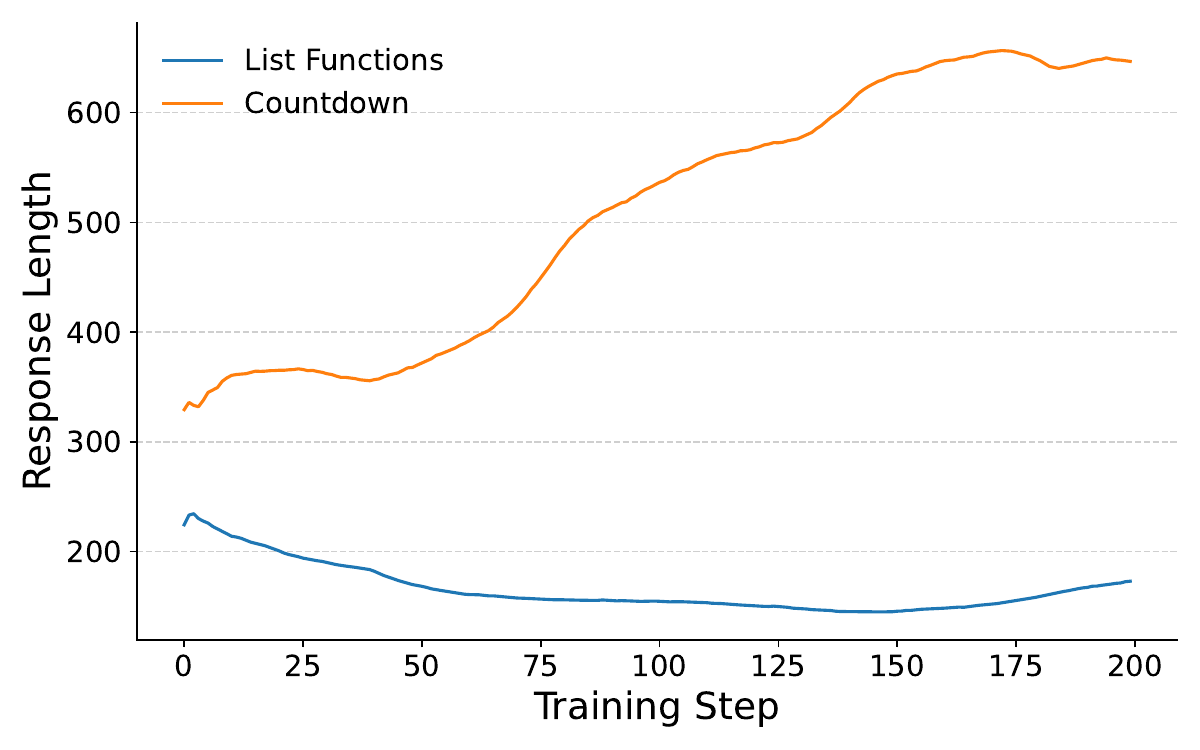}
  \caption{Response length comparison between List Functions and Countdown. Divergent Length Scaling in Countdown vs. List Functions. The chart shows that as training progresses, the model develops longer, more elaborate responses for Countdown tasks, whereas it becomes more brief and direct when performing List Functions.}
\end{figure}

\end{document}